# A New Hip Fracture Risk Index Derived from FEA-Computed Proximal Femur Fracture Loads and Energies-to-Failure

Xuewei Cao[1,#], Joyce H. Keyak[2,#], Sigurdur Sigurdsson[3], Chen Zhao[4], Weihua Zhou[4], Anqi Liu[5], Thomas Lang[6], Hong-Wen Deng[5], Vilmundur Gudnason[3,7,*], Qiuying Sha[1,*]

[1] Department of Mathematical Sciences, Michigan Technological University, Houghton, Michigan, USA

[2] Department of Radiological Sciences, Department of Biomedical Engineering, and Department of Mechanical and Aerospace Engineering, University of California, Irvine, CA, USA

[3] Icelandic Heart Association Research Institute, Kópavogur, Iceland

[4] Department of Applied Computing, Michigan Technological University, Houghton, MI, USA

[5] Center for Bioinformatics and Genomics, School of Medicine, Tulane University, New Orleans, LA, USA

[6] Department of Radiology and Biomedical Imaging, University of California, San Francisco, CA, USA

[7] University of Iceland, Reykjavik, Iceland

[#] Both authors contributed equally

[*] Corresponding authors:

Qiuying Sha, Department of Mathematical Sciences, Michigan Technological University, Houghton, Michigan 49931, USA. E-mail: qsha@mtu.edu

Vilmundur Gudnason, Icelandic Heart Association Research Institute, Kópavogur, Iceland. E-mail: v.gudnason@hjarta.is

## Abstract

Hip fracture risk assessment is an important but challenging task. Quantitative CT-based patient-specific finite element (FE) analysis (FEA) incorporates bone geometry and the three-dimensional distribution of bone density in the proximal femur to compute the force (fracture load) and energy necessary to break the proximal femur in a particular loading condition. Although the fracture loads and energies-to-failure for different loading conditions are individually associated with incident hip fracture, and are mutually correlated, they each provide different structural information about the proximal femur that can influence a subject's overall fracture risk. To obtain a more robust measure of fracture risk, we used principal component analysis (PCA) to develop a global FEA-computed fracture risk index that incorporates the FEA-computed yield and ultimate failure loads and energies-to-failure in four loading conditions (single-limb stance and impact from a fall onto the posterior, posterolateral, and lateral aspects of the greater trochanter) of 110 hip fracture subjects and 235 age- and sex-matched control subjects from the AGES-Reykjavik study. We found that the first principal component (PC1) of the FE parameters was the only significant predictor of hip fracture (p-value < 0.001); we refer to PC1 as the global FEA-computed fracture risk index. To evaluate this fracture risk index, we considered the areal bone mineral density (aBMD) and six covariates including age, sex, height, weight, health status, and bone medication status. Using a logistic regression model, we determined if prediction performance for hip fracture using PC1 differed from that using FE parameters combined by stratified random resampling with respect to hip fracture status. We also compared the performance of predicting hip fracture by the logistic regression model including PC1 with the FRAX model. The results showed that the average of the area under the receiver operating characteristic curve (AUC) using PC1 (0.776) was always higher than that using all FE parameters combined (0.737) in the male subjects (p-value < 0.001). The AUC of PC1 and AUC of the FE parameters combined were not significantly different in the female subjects (p-value = 0.211) or in all subjects (p-value = 0.159). The AUC values using PC1 (0.754 in all subjects, 0.825 in males and 0.71 in females) were greater than the respective values using FRAX (0.651 in the whole sample, 0.705 in males and 0.623 in females) with p-value < 0.01. Therefore, the global FEA-computed fracture risk index based on PCA includes information about hip fracture incidence beyond that of FRAX.

## Introduction

Osteoporosis is among the most common and costly metabolic bone diseases [1], in which the density and quality of bone are reduced. It is characterized by excessive skeletal fragility and susceptibility to low-trauma fracture among the elderly [2-4], causing bones to become weak and brittle and greatly increasing the risk of fracture [5,6]. The incidence and prevalence of osteoporosis increases with age and is related to many factors, such as gender, weight, height, and medical/medication history [7,8]. The increasingly elderly population and the rise in fracture incidence have made osteoporosis a major public health issue in the U.S. and around the world. Osteoporosis affects about 25% of women aged ≥ 65 and about 5% of men aged ≥ 65 [9]. The economic burden of osteoporosis has been estimated at between $17 billion and $20.3 billion in the US alone (2020 data) [10].

Although osteoporosis can affect any bone in the human body, osteoporotic fractures of the proximal femur are the most devastating outcome of the disease, often signaling an end to independent living in the functional elderly. Each year over 300,000 older people in the U.S.—many of those 65 and older—are hospitalized for hip fracture. 5-10% of patients experience a recurrent hip fracture [11]. Hip fractures are invariably associated with chronic pain, reduced mobility, disability, and an increasing degree of dependence [12]. Hip fractures cause the most morbidity among all fractures; reported mortality rates are up to 20-24% in the first year after a hip fracture [13,14], and a greater risk of dying may persist for at least 5 years afterwards [15]. Loss of function and independence among survivors are profound; 40% are unable to walk independently and 60% require assistance a year later [16]. Because of these losses, 33% are totally dependent or in a nursing home in the year following a hip fracture [17,18]. Less than half of those who survive the hip fracture regain their previous level of function [12]. Fracture risk assessment and risk stratification through screening are necessary to reduce the incidence of hip fracture.

Areal bone mineral density (aBMD) of the proximal femur assessed by dual energy X-ray absorptiometry (DXA) is the accepted clinical parameter for the diagnosis of osteoporosis of the hip [19-21]. The aBMD is compared with that of a reference group to obtain the number of standard deviations from the mean aBMD of a gender- and ethnicity-matched healthy population, i.e., the T-score [22,23]. Although low DXA aBMD is associated with bone weakness and fragility fracture [24], DXA is a 2D-projection technique that poorly accounts for bone geometry and size, and it cannot separately evaluate the cortical and trabecular bone [25]; all of these factors influence the integrity of the proximal femur and, therefore, the risk of fracture. Thus, DXA aBMD provides limited information about skeletal factors on fracture risk.

Quantitative CT (QCT) imaging is one of the most powerful methods for assessing bone quality in the proximal femur; after three-dimensional (3D) segmentation of the proximal femur, features such as volumetric bone mineral density (BMD) of the cortical and trabecular bone and bone geometry can be computed. Therefore, QCT is better than DXA for evaluating fracture risk [26]. Patient-specific finite element analysis (FEA) from QCT images incorporates bone geometry, cortical thickness, and the three-dimensional distribution of bone density in the proximal femur to compute the force (fracture load) necessary to break the proximal femur in a particular loading condition. The fracture load can be defined as the force at the onset of fracture (the yield strength) or the maximum force the proximal femur withstands before complete fracture (the ultimate strength or load capacity), and are calculated using FEA with linear or nonlinear material properties, respectively. FEA-computed fracture loads are the most robust measure of proximal femoral structural integrity and, therefore, hip fracture risk [27,28]. In particular, the ultimate strength or load capacity of the proximal femur which is computed with nonlinear FEA is associated with incident hip fracture in men and women, and in men even after controlling for aBMD [8].

Finite element models compute the force necessary to break the proximal femur when forces and other boundary conditions are applied to reflect a specific loading condition, such as single-limb stance during walking or impact onto the greater trochanter from a fall in a specific direction [8]. Although the FEA-computed yield or ultimate strengths for different loading conditions are individually associated with incident hip fracture, and are mutually correlated, each yield and ultimate strength for each loading condition provides different structural information about the proximal femur that contributes to a subject's overall hip fracture risk. The energy transferred to the proximal femur prior to reaching ultimate failure in a particular loading condition may also be associated with fracture risk. Therefore, in this study we used principal component analysis (PCA) to examine a combination of the FEA-computed yield and ultimate strengths and energies-to-failure in various loading conditions to obtain a more robust measure of fracture risk than individual FEA-computed strengths. Principal component analysis is a commonly used method to reduce the dimensionality of data by selecting the most important features that capture the most information [29]. PCA can also speed up the algorithm for prediction by getting rid of correlated variables which do not contribute to decision making [30]. Therefore, we employed PCA to develop a global FEA-computed hip fracture risk index based on the FE model results of 110 hip fracture subjects and 235 age- and sex-matched control subjects in a subset of the AGES-Reykjavik study [31]. PCA is one of the most widely used dimension reduction techniques to transform a large number of variables into a smaller number of variables by identifying the correlations and patterns, while preserving most of the valuable information. The objectives of this study were (1) to construct a global FEA-computed fracture risk index derived from the most important FE parameters of fracture risk based on PCA that is associated with incident hip fracture; (2) to determine if the global FEA-computed fracture risk index, after adjusting for aBMD and covariates, can predict hip fracture better than the FE yield strength, ultimate failure load and energy-to-failure in a single loading condition or their combination in four different loading conditions; and (3) to compare the predictive performance of the global FEA-computed fracture risk index with FRAX.

## Methods

*Subjects*

In this study, we used 110 hip fracture subjects and 235 age- and sex-matched control subjects from the Age Gene/Environment Susceptibility (AGES) Reykjavik study [8,31]. The AGES-Reykjavik study is an ongoing population-based study which contains the baseline QCT scans of subjects who had no metal implants at the level of the hip [31]. Subjects were followed for 4 to 7 years, through November 15, 2009. Forty-two males and 68 females suffered hip fractures during the follow-up period. Meanwhile, 92 male and 143 female control subjects were selected from a pool of age- and sex-matched subjects. The age of the male fracture group was from 71 years to 93 years with a standard deviation (SD) = 5.6 years, and that of the female fracture group was from 67 years to 93 years with SD = 5.9 years (Table 1). Accordingly, the age range of the control male group was from 70 years to 90 years with SD = 5.2 years and that of the control female group was from 67 years to 92 years with SD = 5.7 years. The average heights were 174.6 cm and 174.9 cm for male fracture and control subjects, respectively. The average heights were 159.6 cm and 159.2 cm for female fracture and control subjects, respectively.

*FEA-computed parameters*

The FE models simulated mechanical testing of the femur in which displacement was incrementally applied to the femoral head [8,32]. The computed reaction force on the femoral head initially increased, reached a peak value (the fracture load), and then decreased. To achieve this mechanical behavior, the FE models employed heterogeneous isotropic elastic moduli, yield strengths, and nonlinear post-yield properties. These properties were computed from the calibrated QCT density ($\rho_{CHA}$, g/cm$^3$) of each voxel in an element, which was then used to compute the ash density ($\rho_{ash}$, g/cm$^3$) ($\rho_{ash}$ = 0.0633 + 0.887 $\rho_{CHA}$), and $\rho_{ash}$ was used to compute mechanical properties. Each linear hexahedral finite element measured 3 mm on a side and the mechanical properties of the element were computed by averaging the values of each property over all voxels in the element, while accounting for the volume fraction of each voxel within the element. Together, these mechanical properties described an idealized density-dependent nonlinear stress-strain curve for each element [8,32]. Material yield was defined to occur when the von Mises stress exceeded the yield strength of the element. After yield, the plastic flow was modeled assuming a plastic strain-rate vector normal to the von Mises yield surface and isotropic hardening/softening. Displacement was applied incrementally to the femoral head, and the reaction force on the femoral head was computed at each increment as the distal end of the model was fully constrained. For the fall models, the surface of the greater trochanter opposite the loaded surface of the femoral head was constrained in the direction of the displacements while allowing motion transversely.

Based on our FE modeling method from the QCT data [8, 32-36], twelve FE parameters were evaluated for each subject: yield strength, ultimate failure load, and energy-to-failure calculated during single-limb stance and impact from a fall onto the posterior, posterolateral, and lateral aspects of the greater trochanter (refer to Table 2 for the variable abbreviations). The yield strength was defined as the load at which the von Mises stress in 15 contiguous finite elements exceeded the yield strength for the element. Ultimate failure load was defined the maximum FE-computed force on the femoral head. Energy-to-failure was defined as the area under the force versus displacement curve up to the ultimate failure load. The Pearson correlation coefficient was applied to measure the linear correlation among 12 FE parameters (Figure 1). Although fracture parameters were inherently correlated, the fracture load for each loading condition provided different structural information about the proximal femur that contributes to a subject's overall fracture risk.

*Statistical analysis*

To identify which FE parameters were the most important determinants of fracture risk, multiple linear regression analysis was performed with each FE parameter serving as the dependent variable [8]. Based on our previous study [8], fracture status and the demographic parameters, age, sex, height, and weight were considered as candidate independent variables. To select the most important independent variables in the multiple linear regression, we first performed a simple linear regression model to test the association between each of the candidate independent variables and each of the 12 FE parameters. In the multiple linear regression, we only retained the independent variables with p-value < 0.1. Interactions between fracture status and demographic parameters were also considered as independent variables. Also, if an interaction term was retained, the individual independent variables making up that interaction were retained, regardless of the p-value for the individual independent variable. The multiple linear regression analyses were performed for each of the FE parameters accounting for the retained independent variables. In all of these analyses, FE parameters, age, height, and weight were standardized by subtracting the mean and dividing by the SD of the pooled data.

*Principal component analysis and hip fracture prediction*

The FE parameters for each loading condition provided different structural information about the proximal femur that contribute to a subject's overall fracture risk. Therefore, to obtain a more robust measure of fracture risk, we investigated to use principal component analysis (PCA) to develop a global FEA-computed risk index based on the FE parameters which were mutually correlated. In addition to analyzing data from males and females combined (the whole sample), we also applied PCA to the male sample and female sample, separately. We used a logistic regression model to test the association between hip fracture status and each of the principal components. Let $FX$ be the fracture status, where $FX = 1$ if the subject suffered a hip fracture and $FX = 0$ otherwise. For the $j^{th}$ principal component, the logistic regression model was expressed by $logit(Pr(FX)) = \beta_{0j} + \beta_{1j} PC_j$. The top principal components were retained with p-value < 0.05. To evaluate the hip fracture prediction performance using the global FEA-computed fracture risk index, we considered six covariates which contain four demographic parameters, age, sex, height, and weight, and two clinical parameters, health status (HEALSTAT; excellent, very good, good, fair, or poor) and bone medication status (BMDMED; yes or no). Meanwhile, we also consider the CT-derived total femur areal bone mineral density (aBMD$_{CT}$) which was calculated from the existing QCT scans of the AGES-Reykjavik cohort. In our previous study, aBMD$_{CT}$ has been reported to be correlated with DXA total femur aBMD with $r = 0.935$ [27]. The area under the receiver operating characteristic (ROC) curve (AUC) indicated the predictive performance for each of the classification models, where ROC curve showed the relationship between true positive rate and false positive rate. A larger AUC indicated better the performance of the model at distinguishing between the positive and negative classes. We divided the data based on the incidence of hip fracture into a training set (80% of subjects) and test set (20% of subjects) and analyzed male subjects separately from female subjects as well as analyzing the whole sample. To choose the best predictive model based on the training set, we used three linear classification models, including logistic regression (Logistic), linear discriminant analysis (LDA), and partial least squares analysis (PLS), along with six nonlinear classification models, random forest (RF), quadratic discriminant analysis (QDA), mixture discriminant analysis (MDA), neural networks (NNET), multivariate adaptive regression splines (MARS), and K-nearest neighbors (KNN) [38]. In the training set, we used stratified leave-one-group-out cross-validations (LGOCVs), repeating this procedure 25 times. For each LGOCV, we used 75% of the data to build the classification models and 25% of the data to predict and calculate the AUC. After choosing the best models, we built those models using the entire training set and predicted the fracture status based on the test set. To compare the performance for predicting hip fracture using (a) the global FEA-computed fracture risk index, $aBMD_{CT}$ and covariates, (b) the FE parameters, $aBMD_{CT}$ and covariates, and (c) $aBMD_{CT}$ and covariates, we performed stratified resampling 1000 times. Then we applied a one-sided Student's t-test to compare the resampled AUCs since we were interested in whether the predicative performance of one model was significantly better than the other model.

*Comparison with FRAX*

The fracture risk assessment (FRAX) tool (https://www.sheffield.ac.uk/FRAX/), released in 2008 by the University of Sheffield, computes the individualized 10-year probability of osteoporotic fracture and hip fracture risk [39]. We used the FRAX tool based on the Iceland population. The inputs for FRAX are age, sex, weight, height, previous fracture, parent fractured hip, smoking status, use of oral glucocorticoids, rheumatoid arthritis, and secondary osteoporosis, and alcohol intake. In FRAX, there is an optional input, DXA total femur aBMD. In the AGES-Reykjavik study, there is no DXA total femur aBMD, but there is a CT-derived aBMD, $aBMD_{CT}$. In our previous study comparing $aBMD_{CT}$ with actual aBMD from DXA (Lunar Prodigy, GE Medical Systems, Milwaukee, WI), a strong linear relationship between these two measures was reported (DXA aBMD ($g/cm^2$) = 0.924* $aBMD_{CT}$ ($g/cm^2$) + 0.137; $r = 0.935$; standard error of the estimate (SEE) = 0.046 ($g/cm^2$)) [27]. Therefore, we used this relationship to estimate the DXA-equivalent aBMD based on the $aBMD_{CT}$. We used one-sided DeLong's test to determine if two ROC curves, FRAX and the prediction model using the global FEA-computed fracture risk index, were significantly different.

## Results

Within the male sample, the fracture and control groups were not significantly different with respect to age, height, weight, and the three energies-to-failure in the fall loading conditions (PLenergy, Penergy, Lenergy) at the time of the CT scan (p-value > 0.120; Table 1). However, the remaining nine FE parameters were significantly lower in each fracture group than in the respective control group (p-value < 0.001). In contrast, within the female sample, the fracture and control groups were significantly different with respect to weight (p-value = 0.036). As for the male group, age, height, and the three energies-to-failure in the fall loading conditions in the female group (PLenergy, Penergy, Lenergy) were not significantly different between fracture and control groups (p-value > 0.513). Similarly to the male group, the remaining nine FE parameters were also significantly lower in each fracture group than in the respective control group (Lu, p-value = 0.007; all others, p-value < 0.001; Table 1).

For the multiple linear regression analyses (Table 3), after controlling for four demographic parameters (age, sex, height, and weight) and interactions, the FE parameters except for Pu, and energies-to-failure in the fall loading conditions were associated with hip fracture (p-value < 0.1). However, Pu was significantly lower in each fracture group than in the respective control group when not controlling for demographic parameters and interactions. (p-value < 0.001) (Table 1) and had the higher $R^2 = 0.5217$ (Table 3). Therefore, we considered nine hip fracture-related FE parameters (Sy, Su, Senergy, Py, Pu, PLy, PLu, Ly, and Lu) in the following analysis. PCA was applied to these nine FE parameters, which were inherently highly correlated (Figure 1). The proportions of variance explained by the first principal component (PC1) were 83.31%, 78.23%, and 80.65% for the whole sample, male sample, and female sample, respectively. We found that PC1 of the FE parameters was the only significant predictor for hip fracture (p-value < 0.001 in the whole sample, male sample, and female sample). Therefore, we referred to PC1 as the global FEA computed fracture risk index. Using the LGOCV, we found that the performance of using PC1 along with aBMD$_{CT}$ and covariates, or the nine FE parameters combined along with aBMD$_{CT}$ and covariates, were better than that of only using aBMD$_{CT}$ and covariates to predict hip fracture (Table 4). In particular, we observed the superior predictive performance within the whole sample and male sample of PC1 compared with FE parameters combined for all nine classification models; however, within the female sample, FE parameters combined had greater AUC than PC1. Logistic regression and PLS had the greatest AUCs among nine models (Table 4), and were chosen as the two best models. For PLS and Logistic based on stratified resampling (Figure 2 and Figure 3), in the whole sample, Logistic performed better than the PLS by using PC1 and FE parameters combined along with aBMD$_{CT}$ and covariates, which were also better than Logistic using aBMD$_{CT}$ and covariates (p-value < 0.001). In contrast, within the male sample, PLS using PC1 and FE parameters combined with aBMD$_{CT}$ and covariates was better than PLS using aBMD$_{CT}$ and covariates (p-value < 0.001). In particular, we observed the superior predictive performance in male sample of PC1 compared with the predictive performance of each FE parameter based on PLS (p-value = 0.02 for Ly; all others, p-value < 0.01) (Figure 3). For the female sample, all of the AUCs of PLS using one of the FE parameters were greater than the corresponding AUC using PC1 with p-value > 0.1, except for Ly (p-value = 0.0145).

Finally, to compare the performance of hip fracture prediction using PC1 plus aBMD$_{CT}$ and covariates with FRAX, we used the logistic regression model adding the information of PC1 along with aBMD$_{CT}$ and covariates for three cohorts (whole, male, female). The corrected CT-derived aBMD ranged from 0.137 (g/cm$^2$) to 1.148 (g/cm$^2$) (mean ± SD: 0.597 ± 0.118 (g/cm$^2$)). Figure 4 showed that the AUC using PC1 with aBMD$_{CT}$ and covariates (0.754 in all sample, 0.825 in male and 0.71 in female) was greater than that using FRAX (0.651 in the whole sample, 0.705 in male and 0.623 in female). Thus, the global FEA computed fracture risk index based on PCA includes information about hip fracture incidence beyond that of FRAX (p-value < 0.01) (Figure 4).

## Discussion

This is the first study to construct a global FEA-computed risk index by principal component analysis based on multiple fracture-related FEA-computed fracture loads and energies under different loading conditions. The global FEA-computed fracture risk index, after adjusting for aBMD$_{CT}$ and covariates, predicted hip fracture better than each individual FE parameter (yield strength, ultimate failure load, and energy-to-failure) and also better than the FE parameters combined in the whole sample and the male sample but not the female sample (Figures 2-3, Table 4). Meanwhile, predicting fracture in the female sample was inherently more difficult than predicting fracture in the male sample because the difference between female fracture and control subjects is much smaller than that for males. In our previous work [8], the age-matched design of this study enhanced our ability to explore gender differences in proximal femoral strength and incident hip fracture as a function of age. In particular, our cross-sectional analysis of age-related FE parameter (Su, PLu, Pu, and Lu) loss by gender and fracture status may explain why proximal femoral strength was strongly associated with incident hip fracture in men but much less so in women (Table 3).

Although the FE parameters are individually associated with incident hip fracture, and are mutually highly correlated (Figure 1), they each provide different structural information about the proximal femur that can influence a subject's overall fracture risk. The superior performance, in both men and women, of the assessment of hip fracture risk by using the global FEA-computed risk index and FE parameters combined along with aBMD$_{CT}$ and covariates compared with aBMD$_{CT}$ and covariates is not surprising. The CT-derived total femur aBMD was considered, which has a strong correlation with aBMD from DXA ($r = 0.935$) [27]. The predictive performance by incorporating information from FE parameters were better than only using aBMD$_{CT}$ and covariates in the male sample and female sample (Figure 2, Table 4), implying that the FE parameters can provide additional information in the assessment of hip fracture risk by incorporating bone geometry, cortical thickness, and the three-dimensional distribution of bone density in the proximal femur. Meanwhile, the global FEA-computed risk index plus the aBMD from DXA was better than the aBMD only (Figure 2, Table 4). Although aBMD from DXA correlated with bone weakness and fragility fracture [24], DXA is a 2D-projection technique that poorly accounts for bone geometry and size and is a poor predictor of hip fracture in subjects with osteopenia (T-scores between -1 and -2.5). Thus, aBMD provides limited information about skeletal factors on fracture risk. However, with the emergence of QCT scan-based FE modeling, better estimates of proximal femoral strength have become possible [40]. Our study employed three-dimensional (3-D) FE models from QCT scans, which explicitly represent the 3-D geometry and distribution of material properties that make each femur structurally and mechanically unique and is therefore more robust than two-dimensional (2-D) models from DXA. Principles of physics dictate that hip fracture occurs when an excessive force is applied to the proximal femur, i.e., when the applied force exceeds the force that the proximal femur can support. This force, which varies with the type of loading and force direction, is known as the proximal femoral strength, fracture load [8,34], or load capacity [32] computed using patient-specific FEA. Patient-specific FEA-computed fracture loads and energies are the most robust measures of proximal femoral structural integrity and, therefore, benefitted for evaluating the hip fracture risk.

This global risk index improved assessment of hip fracture risk beyond the commonly used method for evaluating hip fracture risk, namely FRAX (Figure 4), using all sample, the male sample, and the female sample. The FRAX model failed to capture all skeletal determinants of bone strength that tend to be independent of BMD [41,42]. As FRAX only includes BMD from DXA as an input, it may loss information to predict the hip fracture risk. The model may be clinically more appropriate to use the FE parameter from QCT scans in assessing the hip fracture risk of subjects. Although our study has a number of important advantages, such as the age- and sex-matched case-control prospective design and analysis multiple fall loading conditions, the AUC for FRAX of the AGE-Reykjavik cohort was calculated based on the incidence of hip fracture using the estimated 10-year probabilities of hip fracture. Our method was evaluated using the fractures that occurred over 4 to 7 years, which is a limitation to compared with 10-year probabilities of hip fracture from FRAX. However, this study has important clinical implications for the use of 3-D FE models to improve assessment of hip fracture risk, particularly in the high-risk cohort of elderly subjects admitted to hospital following falls. This study identified the clear need for flexibility in using appropriate hip fracture risk calculators to improve healthcare delivery.

## Conclusions

In summary, FEA-computed fracture loads and energies were associated with incident hip fracture in most of loading conditions that were examined, and the global FEA-computed fracture risk index that was investigated by principal component analysis increased hip fracture risk prediction accuracy in the male sample more than that in the female sample. The global FEA-computed fracture risk index was most strongly associated with incident hip fracture in men after accounting for aBMD from DXA and other clinical and demographic parameters. Thus, FE parameters from 3-D FE models includes information about hip fracture beyond that of aBMD from 2-D models and beyond that of FRAX, especially in the male sample. The significance and complexity of these findings, particularly with respect to sex and age effects, indicate that additional studies of FE modeling for hip fracture risk assessment are likely to enhance our understanding of this significant public health problem.

## CONFLICT OF INTERESTS

The authors declare that there is no conflict of interests.

## ACKNOWLEDGMENTS

This study was supported by NIH/NIA R01AG028832 and NIH/NIAMS R01AR46197. The Age, Gene/Environment Susceptibility Reykjavik Study is funded by NIH contract N01-AG-12100, the NIA Intramural Research Program, Hjartavernd (the Icelandic Heart Association), and the Althingi (the Icelandic Parliament). The study was approved by the Icelandic National Bioethics Committee, (VSN: 00-063) and the Data Protection Authority. The researchers are indebted to the participants for their willingness to participate in the study. HWD was partially supported by U19 AG055373 and R01 AR069055. XC was funded by the Michigan Technological University Health Research Institute Fellowship program and the Portage Health Foundation Graduate Assistantship.

**Table 1**. Descriptive statistics for subjects with incident hip fracture and the age- and sex-matched controls. Statistical significance of differences between cases and controls was established using Student's t-test.

| Sex | Measure | Controls | | | | | | Cases | | | | | | p-value |
|---|---|---|---|---|---|---|---|---|---|---|---|---|---|---|
| | | N | Min | Max | Median | Average | SD | N | Min | Max | Median | Average | SD | |
| Male | AGE | 92 | 70 | 90 | 80 | 79.7 | 5.2 | 42 | 71 | 93 | 81 | 80.5 | 5.8 | 0.432 |
| | HEIGHT | 92 | 162.4 | 190.5 | 174.2 | 175 | 6.6 | 42 | 159.2 | 187.1 | 174 | 174.9 | 5.4 | 0.941 |
| | WEIGHT | 92 | 52.4 | 135 | 82 | 82.8 | 14.9 | 42 | 50.3 | 111.3 | 77.5 | 78.7 | 13.5 | 0.120 |
| | PLy | 92 | 638 | 4257 | 1626 | 1704 | 673.9 | 42 | 276 | 2254 | 1202.5 | 1202.4 | 465.1 | <0.001 |
| | PLu | 92 | 2408 | 5499 | 3733.5 | 3838.6 | 570.2 | 42 | 1723 | 4269 | 3253.5 | 3311.1 | 529.5 | <0.001 |
| | Sy | 92 | 2043 | 13990 | 4697.5 | 5010.5 | 1912.5 | 42 | 1261 | 6824 | 3380.5 | 3524.4 | 1219.2 | <0.001 |
| | Su | 92 | 4772 | 21784 | 9992.5 | 10628.1 | 2967.8 | 42 | 3123 | 12898 | 7980.5 | 7980.2 | 2017.2 | <0.001 |
| | Py | 92 | 631 | 4123 | 1699 | 1821.3 | 655.7 | 42 | 404 | 2014 | 1295 | 1294.6 | 415.9 | <0.001 |
| | Pu | 92 | 2149 | 5078 | 3322 | 3322.9 | 456.1 | 42 | 1766 | 3964 | 3013 | 2954.9 | 426.3 | <0.001 |
| | Ly | 92 | 1028 | 5393 | 2063 | 2141.5 | 745.3 | 42 | 441 | 2957 | 1427 | 1476.1 | 507.8 | <0.001 |
| | Lu | 92 | 2344 | 6619 | 4299.5 | 4315.2 | 743.9 | 42 | 1913 | 4794 | 3776 | 3669.1 | 598.1 | <0.001 |
| | Senergy | 92 | 287.9 | 2591 | 902.9 | 963.3 | 425.5 | 42 | 189.2 | 1525.2 | 641.5 | 673 | 282.2 | <0.001 |
| | PLenergy | 92 | 233.8 | 991.9 | 471.4 | 496 | 139.8 | 42 | 172.5 | 1275.1 | 449.3 | 483.5 | 176.4 | 0.686 |
| | Penergy | 92 | 189.9 | 788.5 | 406.4 | 423.1 | 137.3 | 42 | 218.3 | 609.7 | 408.3 | 420.3 | 97.7 | 0.890 |
| | Lenergy | 92 | 89.6 | 944.2 | 477.2 | 475.1 | 161.8 | 42 | 123.8 | 899.6 | 440.2 | 452.2 | 174.1 | 0.474 |
| Female | AGE | 143 | 67 | 92 | 79 | 79.2 | 5.7 | 68 | 67 | 93 | 79.5 | 79.7 | 6.1 | 0.544 |
| | HEIGHT | 143 | 139.2 | 172.9 | 159.5 | 159.3 | 5.6 | 68 | 145.1 | 173.4 | 159.5 | 159.3 | 6.1 | 0.995 |
| | WEIGHT | 143 | 37.2 | 112 | 66.4 | 68.7 | 13.7 | 68 | 39.2 | 111.2 | 63.4 | 64.2 | 15 | 0.036 |
| | PLy | 143 | 314 | 3071 | 987 | 1093.3 | 476.5 | 68 | 353 | 1945 | 780 | 850 | 336 | <0.001 |
| | PLu | 143 | 1816 | 4535 | 2837 | 2963.9 | 551.2 | 68 | 1700 | 4419 | 2632 | 2693.3 | 440 | <0.001 |
| | Sy | 143 | 1272 | 9446 | 2924 | 3362.6 | 1543.1 | 68 | 1300 | 5913 | 2528 | 2680.2 | 891.6 | <0.001 |
| | Su | 143 | 3611 | 16330 | 6775 | 7256.1 | 2335.8 | 68 | 3085 | 10675 | 5831.5 | 6045.3 | 1609.2 | <0.001 |
| | Py | 143 | 525 | 2733 | 1154 | 1247.8 | 468.9 | 68 | 490 | 2049 | 928 | 1006.9 | 341.6 | <0.001 |
| | Pu | 143 | 1933 | 3893 | 2539 | 2641.6 | 399.4 | 68 | 1610 | 3150 | 2373 | 2425.3 | 349 | <0.001 |
| | Ly | 143 | 469 | 3736 | 1319 | 1432.2 | 585.8 | 68 | 570 | 2532 | 1004 | 1074.5 | 382 | <0.001 |
| | Lu | 143 | 1941 | 5386 | 3203 | 3256.7 | 650.6 | 68 | 1945 | 4896 | 2989.5 | 3019.9 | 551.4 | 0.007 |
| | Senergy | 143 | 195.2 | 1770.3 | 492.6 | 558.4 | 275.8 | 68 | 196.5 | 1115.5 | 396.7 | 423.3 | 155.1 | <0.001 |
| | PLenergy | 143 | 56.1 | 800.1 | 403.6 | 414.6 | 125.3 | 68 | 118.6 | 912.6 | 393.2 | 412.1 | 139 | 0.899 |
| | Penergy | 143 | 84.9 | 787.1 | 362.9 | 377 | 118.9 | 68 | 158.1 | 606.2 | 360.8 | 367 | 96.5 | 0.513 |
| | Lenergy | 143 | 86.1 | 943.7 | 385.8 | 390.9 | 154.3 | 68 | 119.2 | 775.4 | 384.5 | 382.8 | 147 | 0.714 |

**Table 2.** The abbreviations of FE parameters.

| FE Parameters | Loading Condition | | | |
| --- | --- | --- | --- | --- |
| | Stance | Posterior | Posterolateral | Lateral |
| Yield strength (force at onset of fracture) | Sy | Py | PLy | Ly |
| Ultimate strength (load capacity) | Su | Pu | PLu | Lu |
| Energy to failure | Senergy | Penergy | PLenergy | Lenergy |

**Table 3**. Multiple linear regression results for each of the 12 FE parameters for all subjects, including coefficients of standardized variables, standard errors, and p-values.

| Dependent variable | Regression coefficients for standardized variables (Standard error) p-value | | | | | | | | $R^2$ |
|---|---|---|---|---|---|---|---|---|---|
| | FX | Sex | Age | Weight | Height | Fx:Age | Fx:Sex | Fx:Height | |
| PLy | **-0.301** (0.115) **0.009** | **0.706** (0.158) **<0.001** | **-0.147** (0.055) **0.007** | **0.298** (0.056) **<0.001** | 0.033 (0.107) 0.759 | - | **-0.420** (0.183) **0.022** | - | 0.3773 |
| PLu | **-0.299** (0.101) **0.003** | **0.833** (0.138) **<0.001** | **-0.151** (0.040) **<0.001** | **0.294** (0.047) **<0.001** | **0.130** (0.072) **0.072** | - | **-0.400** (0160) **0.013** | - | 0.5467 |
| Sy | **-0.290** (0.148) **0.050** | **0.607** (0.179) **0.001** | **-0.244** (0.069) **<0.001** | **0.237** (0.057) **<0.001** | 0.097 (0.124) 0.435 | 0.144 (0.115) 0.212 | -0.491 (0.320) 0.127 | 0.028 (0.209) 0.892 | 0.3571 |
| Su | **-0.315** (0.105) **0.003** | **0.651** (0.144) **<0.001** | **-0.183** (0.050) **<0.001** | **0.327** (0.050) **<0.001** | **0.146** (0.075) **0.052** | 0.121 (0.080) 0.134 | **-0.500** (0.167) **0.003** | - | 0.511 |
| Py | **-0.325** (0.115) **0.005** | **0.696** (0.159) **<0.001** | **-0.268** (0.067) **<0.001** | **0.248** (0.057) **<0.001** | 0.060 (0.107) 0.576 | 0.165 (0.106) 0.121 | **-0.493** (0.184) **0.008** | - | 0.3806 |
| Pu | -0.124 (0.137) 0.366 | **0.941** (0.159) **<0.001** | **-0.151** (0.051) **0.003** | **0.243** (0.049) **<0.001** | 0.077 (0.085) 0.364 | **0.234** (0.086) **0.007** | **-0.800** (0.287) **0.006** | **0.332** (0.143) **0.021** | 0.5217 |
| Ly | **-0.417** (0.111) **<0.001** | **0.713** (0.153) **<0.001** | **-0.295** (0.064) **<0.001** | **0.234** (0.054) **<0.001** | 0.069 (0.103) 0.505 | **0.187** (0.102) **0.069** | **-0.437** (0.177) **0.014** | - | 0.4201 |
| Lu | **-0.201** (0.104) **0.054** | **0.780** (0.142) **<0.001** | **-0.183** (0.050) **<0.001** | **0.266** (0.049) **<0.001** | **0.180** (0.074) **0.015** | 0.125 (0.079) 0.117 | **-0.492** (0.165) **0.003** | - | 0.521 |
| Senergy | **-0.299** (0.140) **0.034** | **0.593** (0.170) **0.001** | **-0.168** (0.066) **0.011** | **0.288** (0.054) **<0.001** | **0.203** (0.118) **0.085** | 0.121 (0.110) 0.274 | -0.360 (0.305) 0.239 | -0.053 (0.199) 0.789 | 0.4343 |
| Fenergy | - | 0.112 (0.160) 0.485 | - | **0.111** (0.064) **0.082** | **0.247** (0.117) **0.035** | - | - | - | 0.109 |
| Penergy | -0.089 (0.111) 0.424 | 0.047 (0.181) 0.797 | 0.075 (0.082) 0.359 | -0.007 (0.069) 0.918 | **0.274** (0.131) **0.037** | 0.205 (0.130) 0.116 | - | - | 0.0677 |
| Lenergy | - | 0.190 (0.178) 0.286 | -0.096 (0.066) 0.144 | **0.121** (0.067) **0.072** | 0.157 (0.128) 0.221 | - | - | - | 0.0991 |

*Notes:* The bold-faced value means the variable is significant (p-value < 0.1). "–" indicates that the variable was not included in the model because the p-values would have been greater than 0.1.

**Table 4**. The average AUCs of the classification models using stratified leave-one-group-out cross-validation for the fracture status prediction using aBMD$_{CT}$ and covariates, PC1 plus aBMD$_{CT}$ and covariates, and the nine FE parameters combined plus aBMD$_{CT}$ and covariates, respectively.

| Model | Sample | Average AUC (Standard error) | | |
|---|---|---|---|---|
| | | aBMD$_{CT}$ + Covariates | PC1 + aBMD$_{CT}$ + Covariates | FE parameters combined + aBMD$_{CT}$ + Covariates |
| Logistic | Whole | 0.699 (0.052) | **0.738 (0.043)** | 0.724 (0.041) |
| | Male | 0.727 (0.093) | **0.777 (0.093)** | 0.745 (0.094) |
| | Female | 0.608 (0.115) | 0.623 (0.095) | **0.669 (0.089)** |
| LDA | Whole | 0.696 (0.053) | **0.725 (0.046)** | 0.708 (0.044) |
| | Male | 0.727 (0.090) | **0.758 (0.099)** | 0.751 (0.105) |
| | Female | 0.608 (0.116) | 0.615 (0.110) | **0.659 (0.080)** |
| PLS | Whole | 0.700 (0.049) | **0.737 (0.045)** | 0.731 (0.049) |
| | Male | 0.719 (0.093) | **0.788 (0.113)** | 0.753 (0.117) |
| | Female | 0.617 (0.068) | **0.638 (0.087)** | 0.611 (0.086) |
| RF | Whole | 0.640 (0.045) | 0.684 (0.042) | **0.727 (0.049)** |
| | Male | 0.714 (0.107) | **0.745 (0.086)** | 0.742 (0.072) |
| | Female | 0.555 (0.060) | **0.577 (0.075)** | 0.569 (0.075) |
| QDA | Whole | 0.637 (0.073) | **0.695 (0.056)** | 0.681 (0.052) |
| | Male | 0.690 (0.110) | **0.707 (0.141)** | 0.674 (0.101) |
| | Female | 0.584 (0.085) | 0.622 (0.067) | **0.649 (0.083)** |
| MDA | Whole | 0.668 (0.056) | **0.702 (0.047)** | 0.661 (0.048) |
| | Male | 0.724 (0.078) | **0.719 (0.105)** | 0.664 (0.102) |
| | Female | 0.585 (0.074) | 0.632 (0.066) | **0.645 (0.064)** |
| NNET | Whole | 0.693 (0.061) | **0.731 (0.050)** | 0.719 (0.046) |
| | Male | 0.699 (0.109) | **0.744 (0.110)** | 0.714 (0.093) |
| | Female | 0.632 (0.115) | 0.563 (0.136) | **0.622 (0.099)** |
| MARS | Whole | 0.668 (0.052) | **0.717 (0.039)** | 0.689 (0.054) |
| | Male | 0.573 (0.127) | **0.724 (0.115)** | 0.664 (0.137) |
| | Female | 0.585 (0.086) | 0.558 (0.094) | **0.667 (0.085)** |
| KNN | Whole | 0.607 (0.120) | 0.689 (0.074) | **0.691 (0.065)** |
| | Male | 0.671 (0.128) | **0.744 (0.096)** | 0.681 (0.128) |
| | Female | 0.511 (0.078) | 0.517 (0.109) | **0.554 (0.104)** |

*Notes*: The bold-faced value means the largest AUCs among three comparison models.

**Figure 1**. The correlation coefficients among 12 FE parameters for all subjects.

**Figure 2**. The predictive performance of Logistic and PLS based on stratified resampling. The values over the lines indicated the p-values obtained from the one-sided Student's t-test. The values on top of the boxes indicated the average AUCs by using PC1 along with aBMD$_{CT}$ and covariates (PC1 + aBMD$_{CT}$ + Cov) versus using aBMD$_{CT}$ and covariates (aBMD$_{CT}$ + Cov) or FE parameters combined with aBMD$_{CT}$ and covariates (FE combined + aBMD$_{CT}$ + Cov).

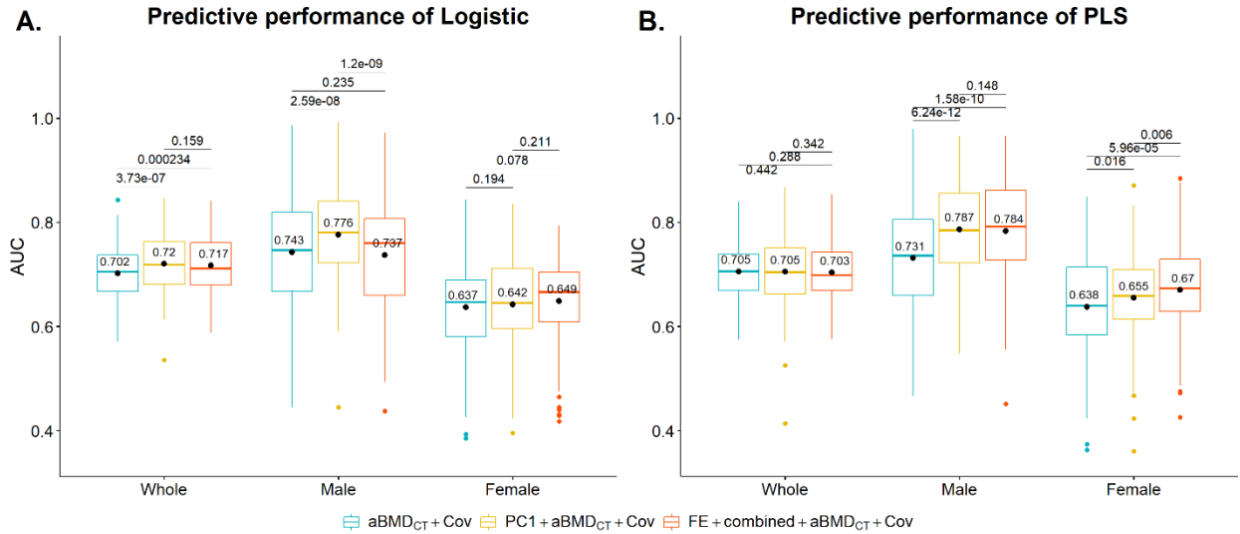

**Figure 3**. The predictive performance of PLS using each FE parameters compared with that of using PC1. The values over the lines indicated the p-values were obtained from the one-sided Student's t-test. The values on top of the boxes indicated the average AUCs by using PC1 along with aBMD$_{CT}$ and covariates (PC1 + aBMD$_{CT}$ + Cov) versus using each of FE parameters along with aBMD$_{CT}$ and covariates (* + aBMD$_{CT}$ + Cov).

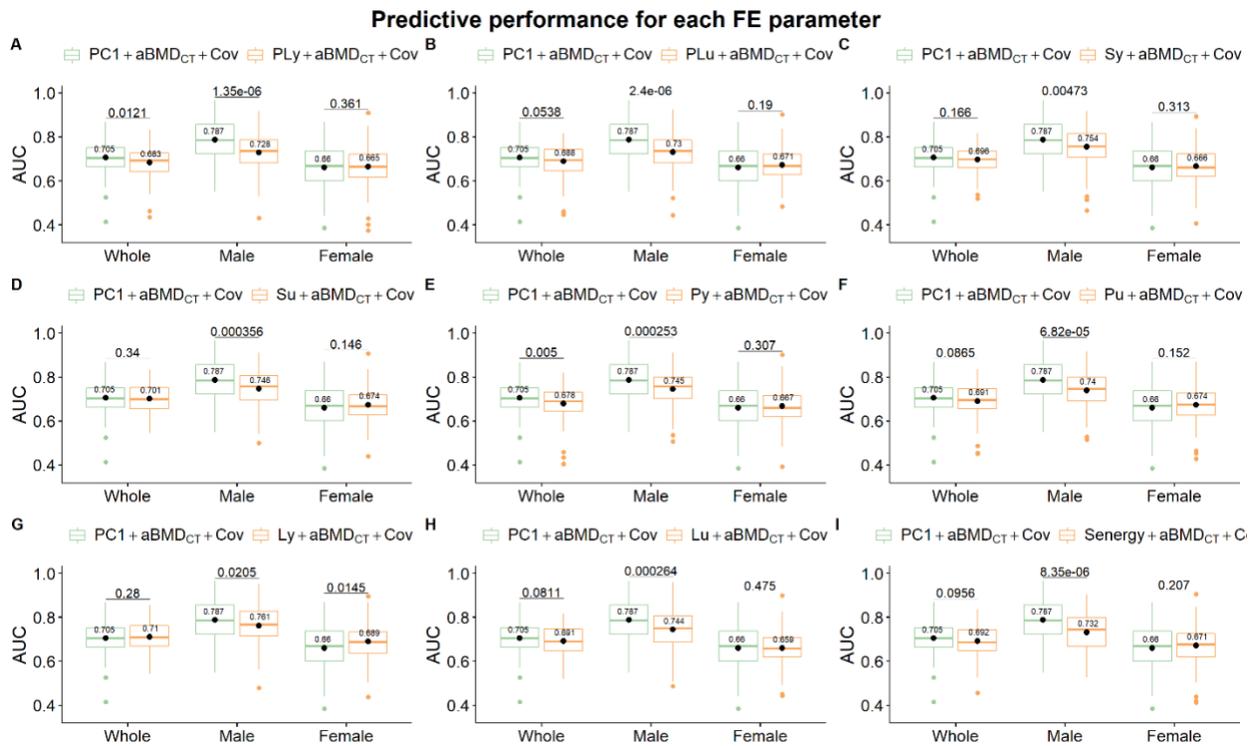

**Figure 4.** The performance of the hip fracture prediction for PC1 and FRAX for the whole sample (left), male sample (middle), and female sample (right).

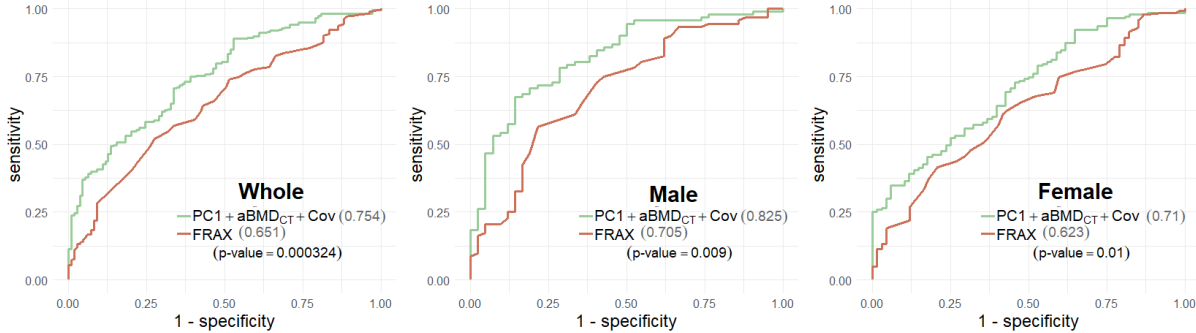